\documentclass{sig-alternate}

\usepackage{hyperref}
\usepackage{amssymb}
\usepackage{amsmath}
\usepackage{multirow}
\usepackage{tabularx}

\begin{document}
%
% --- Author Metadata here ---
\conferenceinfo{Foundations of Digital Games}{???}
%\CopyrightYear{2013} % Allows default copyright year (20XX) to be over-ridden - IF NEED BE.
%\crdata{0-12345-67-8/90/01}  % Allows default copyright data (0-89791-88-6/97/05) to be over-ridden - IF NEED BE.
% --- End of Author Metadata ---

\title{Automatic Playtesting for Game Parameter Tuning via Active Learning}

\numberofauthors{1} 

\author{
\alignauthor
%Alexander Zook and Mark O. Riedl\\
%       \affaddr{School of Interactive Computing, College of Computing}\\
%       \affaddr{Georgia Institute of Technology}\\
%       \affaddr{Atlanta, Georgia, USA}\\
%       \email{\{a.zook, riedl\}@gatech.edu}
%
Alexander Zook, Eric Fruchter and Mark O. Riedl\\
       \affaddr{School of Interactive Computing, College of Computing}\\
       \affaddr{Georgia Institute of Technology}\\
       \affaddr{Atlanta, Georgia, USA}\\
       \email{\{a.zook, efruchter, riedl\}@gatech.edu}
}

\toappear{} % for FDG to fill in

\maketitle
\begin{abstract}
Game designers use human playtesting to gather feedback about game design elements when iteratively improving a game.
Playtesting, however, is expensive: human testers must be recruited, playtest results must be aggregated and interpreted, and changes to game designs must be extrapolated from these results.
Can automated methods reduce this expense?
We show how active learning techniques can formalize and automate a subset of playtesting goals.
Specifically, we focus on the low-level parameter tuning required to balance a game once the mechanics have been chosen.
Through a case study on a shoot-`em-up game we demonstrate the efficacy of active learning to reduce the amount of playtesting needed to choose the optimal set of game parameters for two classes of (formal) design objectives.
This work opens the potential for additional methods to reduce the human burden of performing playtesting for a variety of relevant design concerns.
\end{abstract}

\category{Applied Computing}{Computers in other domains}{Personal computers and PC applications}[Computer games]
%A category including the fourth, optional field follows...
%\category{D.2.8}{Software Engineering}{Metrics}[complexity measures, performance measures]

\terms{Experimentation}


\section{Introduction}

%\todo[inline]{update top part w/FDG info} % TODO!

% motivation
Iterative game design practices emphasize the centrality of playtesting to improve and refine a game's design.
Human playtesters provide valuable feedback on audience reactions to a game.
Playtesting is often claimed to be ``the single most important activity a designer engages in'' \cite{fullerton2008:playcentric}.
Test data informs designers of how real players may react to the game in ways that self-testing, simulations, and design analysis may not.
Playtesting, however, is expensive---developers must recruit players, devise design experiments, collect game play and subjective feedback data, and make design changes to meet design goals.

%  other quotes:
%``You might be thinking: But testing is an expensive process, isn't it?'' p.248

% problem
We ask the question: can we reduce the cost of the playtesting process by automating some of the more mundane aspects of playtesting?
To address this problem we examine a subset of playtesting questions focused on ``parameter tuning.''
Parameter tuning involves making low-level changes to game mechanic settings such as character movement parameters, power-up item effects, or control sensitivity.
Games based on careful timing and reflexes depend on well-tuned parameters, including racing, platforming, shoot-`em-up, and fighting game genres.
Addressing the problem of parameter tuning requires a means to automatically select a set of potentially good parameter settings, test those settings with humans, evaluate the human results, and repeat the process until a pre-defined design goal is achieved.

% contributions
Our primary insight is to model playtesting as a form of active learning (AL).
Active learning \cite{settles2012:al-book} selects among a set of possible inputs to get the best output while minimizing the number of inputs tested.
We define the ``best output'' as a parameter tuning design goal and treat a set of game design parameters as an ``input.''
Minimizing the number of inputs tested minimizes the number of playtests performed, saving human effort and reducing costs.
This paper makes three contributions toward machine--driven playtesting:
\begin{enumerate}
\item Formulating efficient playtesting as an AL problem
\item Defining several common playtesting goals in terms of AL metrics
\item Demonstrating the efficacy of AL to reduce the number of playtests needed to optimize (1) difficulty-related and (2) control-related game parameters in a case study of a shoot-`em-up game
\end{enumerate}

% differences from prior work
We believe machine--driven playtesting is a novel use of machine learning in games.
Unlike prior work on dynamic difficulty and adaptive games we focus on the case of deciding on a fixed design for future use.
Our approach can apply to iterative, online game adjustments to more rapidly converge on the right set of game parameters.
%Further, by learning a model of how design parameters influence playtest results our approach is able to provide designers additional feedback on their ``design space'' at design time.
%A design space can be visualized to understand the tradeoffs among different parameter settings.
Unlike prior work on game design support tools using simulations or model-checking we focus on the problem of efficiently working with a set of human testers.
Our approach complements these tools for early-stage exploration with late-stage refinement.
To the best of our knowledge this work is novel in automatically tuning game controls. %, specifically meeting player preferences.
Controls are an important element of player \textit{interaction}.
Poor controls can obstruct player enjoyment or make a game unplayable, yet have been largely overlooked in AI for game design support.
This work provides a first step to addressing interaction through tuning controls to construct a single control setting that best meets the preferences of many players (e.g. default controls).

% roadmap
In this paper we first compare our approach to related work on game design support.
We next define a set of parameter tuning design goals and relate them to AL methods.
Following this we describe a case study of automated playtesting using a shoot-`em-up game.
After describing the game we present results showing AL reduces playtesting costs using human data collected from an online study.
We conclude with a discussion of the limitations, applications, and potential extensions to this playtesting approach.

%%%%%%%%%%%%%%%%%%%%%%%%%%%%%%%%%%%

\section{Related Work}

% overview
Two research areas are closely related to machine playtesting: offline game design tools and online game adaptation.
Offline game design tools help designers explore possible game designs by defining a high-level space of games through a design language.
%Examples include support for generating or evaluating design in terms of hard constraints (what must be true) or soft optimization (what to achieve to the best degree possible).
Online game adaptation changes game designs in real-time based on player actions or game state.
%Examples include designer authored rules or data-driven player models coupled to optimization techniques.

\subsection{Offline Game Design Tools}
% offline design: model-checking vs simulation
Offline game design tools have evaluated game designs using simulated players and formal model-checking.
Simulation-based tools use sampling techniques to test aspects of a game design for ``playability,'' typically defined as the ability for a player to reach a given goal state with the current design parameters.
Model-checking tools define game mechanics in a logical language to provide hard guarantees on the same kinds of playability tests.

% simulation
Bauer et al. \cite{bauer2013:rrt-generation} and Cook et al. \cite{cook2012:coopcoevo} use sampling methods to evaluate platformer game level playability.
Shaker et al. \cite{shaker2013:ropossum-test} combine a rule-based reasoning approach with simulation to generate content for a physics-based game.
Simulation approaches are valuable when design involves an intractably large space of possible parameters to test and can serve as input to optimization techniques.
%
% model-checking
Model-checking approaches provide guarantees on generated designs having formally defined properties---typically at the cost of being limited to more coarse design parameter decisions.
Smith et al. \cite{smith2013:quantify-play}, Butler et al. \cite{butler2013:progression-tool}, and Horswill and Foged \cite{horswill2012:levelgen} use logic programming and constraint solving to generating levels or sets of levels meeting given design constraints.
Jaffe et al. \cite{jaffe2012:balance} use game-theoretic analysis to understand the effects of game design parameters on competitive game balance.

% differences of playtest vs model-check -> value to designers
Our approach to automated playtesting complements these approaches to high-level early design exploration with low-level optimization of game parameters and tuning.
Further, we focus on a generic technique that applies to cases with human testers in the loop, crucial to tuning game controls or subjective features of games.
Offline design tools currently enable designers to formally define and enforce properties of a game design across all possible player behaviors in a specific game or space of designs.
To date these tools have emphasized techniques for ensuring game designs have desired formal properties that meet designer intent.
%
% note: not about model-checking vs playtesting but expanding offline tools available
Machine--driven playtesting uses players to provide feedback to designers on \textit{expected} player behaviors in a game.
Developing machine--driven playtesting techniques affords designers insight into how human audiences interact with designer intent, complementing an understanding of whether and how a game matches formal criteria.
AL complements formal design techniques by allowing them to more efficiently learn and improve models.

\subsection{Online Game Adaptation}
%% online adaptation: hand-crafted rules vs ML/EC
% hand-crafted rules
Online game adaptation researchers have used both hand-crafted rules and data-driven techniques.
Hunicke and Chapman \cite{hunicke2004:dda} track the average and variance of player damage and inventory levels and use a hand-crafted policy to adjust levels of enemies or powerups. 
Systems by Magerko et al. \cite{magerko2006:isat}, El-Nasr \cite{seifel-nasr2007:mirage}, and Thue et al. \cite{thue2007:storytell-pm} model players as vectors of skills, personality traits, or pre-defined ``player types'' and select content to fit players using hand-crafted rules. % note: erring on side of more for now
Hand-crafted rules allow designers to give fine-tuned details of how to adjust a game toward design goals.
However, designers must fully describe how to change the game and rules are often sensitive to minor game design changes.

% ML / EC
To bypass the brittleness of rules others have used data-driven techniques that optimize game parameters toward design goals.
Hastings et al. \cite{hastings2009:gar}, Shaker et al. \cite{shaker2013:crowdsource-platform-aesthetics}, Liapis et al. \cite{liapis2013:rank-based-interactive-evol} and Yu and Riedl \cite{yu2013:storyeti} model player preferences using neuro-evolutionary or machine learning techniques and optimize the output of these models to select potential game parameters.
Harrison and Roberts \cite{harrison2013:scrabble-retention} optimize player retention and Zook and Riedl \cite{zook2012:tf} optimize game difficulty with similar techniques.

%Both rule-based and data-driven online adaptation approaches target real-time adjustments of a game to known design goals.
%Active learning can enhance simulation approaches by guiding the sampling process toward spaces of parameters likely to be of greater value.
Automated playtesting extends these approaches with principled methods to guide the process of designing hand-crafted rules or optimizing game parameters.
When hand-crafting rules, automated playtesting informs the choice of which rule parameters to use.
When optimizing models learned from data, automated playtesting informs the choice of which next set of parameters to test during the optimization process.
We argue research to date has ignored the problem of reducing ``sample complexity''---the number of data points (human playtests) needed to train a model.
Active learning makes explicit the trade-off in playtesting between ``exploring'' potentially valuable game design settings and ``exploiting'' known good solutions with small changes.
%Active learning complements the above approaches by providing a principled method for reducing the number of playtests needed without changing the fundamental models involved.
Thus, AL can complement online game adaptation by reducing the number of mediocre or bad sets of game parameters players experience before arriving at good parameter settings without changing the underlying models used.

\subsection{Active Learning in Games}

There are other uses of AL in games.
% previous AL approaches
Normoyle et al. \cite{normoyle2012:al-metrics} use AL to recommend sets of useful player metrics to track.
Rafferty et al. \cite{rafferty2012:opt-cog-game} optimize game designs offline to learn the most about player cognition.
Machine--driven playtesting complements prior uses of AL for game design by focusing on efficiently improving designs for player behavior and experience.
We extend these efforts with a wider variety of AL methods while addressing the cost of playtesting.

%%%%%%%%%%%%%%%%%%%%%%%%%%%%%%%%%%

\section{Playtesting as Active Learning}

Our goal is to automate mundane playtesting tasks by efficiently choosing game design parameters for players to test.
Active learning provides a generic set of techniques to perform the playtesting process of choosing a set of design parameters to test toward achieving a design goal.
Playtesting typically involves trade-offs between testing designs that are poorly understood (exploration) and refining designs that are known to be good but need minor changes (exploitation).
Active learning captures this intuition through explicit models of the exploration-exploitation trade-off.
In this section we characterize playtesting in terms of AL.
We provide intuitions behind AL, but full mathematical treatments are available through the references.

Machine--driven playtesting---AL for game design---involves (1) a design model for how a design works, (2) a design goal, and (3) a playtesting strategy for how to choose a new design when seeking the best design for a goal.
Formally, a design model is a \textit{function} that captures the relationship between game design parameters (input) and game metrics (output: e.g. specific player behaviors or subjective responses).
The design goal is an \textit{objective function} specifying what game metrics are desired.
Playtesting strategies choose what design to test next using an \textit{acquisition function} that uses information from the design model and goal to predict and value possible playtest outcomes.

Design model functions come in two forms: regression and classification.
\textit{Regression} models predict how design parameters will change continuous outputs---e.g. how many times a player is hit or how long it takes players to complete a platformer level.
\textit{Classification} models predict changes in discrete outputs---e.g. which of a pair of racing game control settings a player will prefer or which choose-your-own adventure choice a player will make when given a set of options.
Objective functions specify how to value different outputs---e.g. wanting players to be hit a certain number of times or wanting players to agree that one set of controls is good.
Note that many design goals can be formulated as goals for playtesting: the challenge lies in defining a useful metric for measuring these goals through player feedback or in-game behavior.

Acquisition functions differ for regression and classification models.
In the next sections we provide intuitive descriptions of several acquisition functions---references to the relevant mathematical literature are provided.
Our survey of regression models covers key methods along the exploration-exploitation spectrum.
For classification models we cover the most common frameworks for AL with discrete data that are intended to mitigate the impact of highly variable data.

\subsection{Regression Models}
%% regression
Acquisition functions choose which input parameters to test next to most efficiently maximize an objective function.
Acquisition functions vary along a spectrum of exploration---playtesting by picking designs that are least understood---and exploitation---picking designs that are expected to be best.
We consider four acquisition functions for regression models: (1) variance, (2) probability of improvement, (3) expected improvement, and (4) upper-confidence bounds.
These acquisition functions were developed in the field of Bayesian experimental design and apply generally to any regression model with a probabilistic interpretation \cite{brochu2010:thesis, chaloner1995}.
Regression models are useful when design goals fall along a continuous scale; we examine player behavior, specifically how enemy design parameters affect player performance.

Regression acquisition functions include:
\begin{itemize}
\item \textbf{Variance} Exploration by picking the input with greatest output variance \cite{brochu2010:thesis}.
% Picks the input with greatest output uncertainty \cite{brochu2010:thesis}. 
% Corresponds to picking the design that is hardest to predict the outcomes from. 
% Exploration.
%
\item \textbf{Probability of Improvement (PI)} Exploitation by picking the input most likely to have an output that improves over the previous best \cite{brochu2010:thesis}.
% Picks the input that most likely to improves the over the best previous output \cite{brochu2010:thesis}. 
% Corresponds to picking the design most likely to improve over the current best. 
% Exploitation.
%
\item \textbf{Expected Improvement (EI)} Balances exploration and exploitation by picking the input with greatest combined probability and amount of improvement over the previous best \cite{brochu2010:thesis}.
% Picks the input by weighting the probability of output improvement by the amount of improvement \cite{brochu2010:thesis}.
% Corresponds to picking the design with largest expected improvement.
% Balances exploration and exploitation, but more computationally costly.
%
\item \textbf{Upper Confidence Bound (UCB)}
Balances exploration and exploitation by picking the input with greatest combined expected value and uncertainty to gradually narrow the space of inputs \cite{srinivas2010:gp-ucb}.
% Picks the input with combined greatest expected value and uncertainty \cite{srinivas2010:gp-ucb}.
% Corresponds to picking designs that seem high quality but are poorly understood to gradually narrow the space of design to be known good or expected bad but uncertain.
% The approach cited gradually reduces the weighting of uncertainty to help converge on an answer.
% Balances exploration and exploitation.
\end{itemize}

\subsection{Classification Models}
%% classification
Classification models are useful when design goals involve discrete choices; we examine player subjective ratings, specifically studying preference choices when picking between sets of controls.
Classification models are primarily concerned with increasing certainty in predicting outcomes---improving the model of how the design works.
We consider five acquisition functions for classification models: (1) entropy, (2) query-by-bagging (QBB) vote, (3) query-by-bagging (QBB) probability, (4) expected error reduction, and (5) variance reduction.
These acquisition functions have been developed for classification models; several---entropy, QBB probability, and expected error and variance reduction---require probabilistic predictions.
% entropy
% note: technically a subset of uncertainty sampling
% (1) least confident
% (2) margin-based
% (3) entropy
% (4) expected error reduction
% (5) variance reduction

\begin{itemize}
\item \textbf{Entropy} Picks the input with greatest output uncertainty according to entropy---a measure of the amount of information needed to encode a distribution \cite{settles2012:al-book}.
% Picks input with greatest output uncertainty according to entropy---a measure of information needed to encode a distribution \cite{settles2012:al-book}.
% Corresponds to picking designs where player choices are most uncertain.
% Exploration.
%
\item \textbf{Query-By-Bagging (QBB)} Picks the input with most disagreement among copies of a classification model trained on random subsets of known results \cite{settles2012:al-book}.
% Trains multiple copies of a classification model on random subsets of existing input-output data (``bags'') and picks the next input with greatest disagreement among those models \cite{settles2012:al-book}.
\textbf{QBB vote} picks the input with the largest difference between its top two output options \cite{settles2012:al-book}.
\textbf{QBB probability} picks the output with greatest average uncertainty across the models \cite{abe1998:qbb}.
% Corresponds to picking designs with greatest variability in expected outcomes when viewed based on different subsets of playtest data.
% Exploration.
% Two varieties are relevant:
% \begin{description}
% \item[QBB vote] Each model votes on predicted output for each potential input.
% The input with the greatest difference between its top two output options is then picked.
% \item[QBB probability] Each model predicts the probability of each predicted output for each potential input.
% The input with most uncertain averaged output probability is then picked \cite{abe1998:qbb}.
%
\item \textbf{Expected Error Reduction} Picks the input that, if used to train the model, leads to the greatest expected reduction in classification error \cite{settles2012:al-book}.
% Picks the input that, if used to train the model, will give greatest expected reduction in model error when predicting on the remaining inputs.
% Corresponds to picking designs that will most improve prediction of the design outcomes over the design as a whole.
% Balances exploration and exploitation but very computationally expensive.
% The algorithm performs the following steps:
% \begin{enumerate}
% \item Pick a potential input and assign one of the possible outputs to make a ``fake'' training point.
% \item Train a new classification model using the fake point and compute the error over the remaining input points.
% \item Repeat (1) and (2) for all possible outputs and compute the average (``expected'') error across possible outputs.
% \item Repeat (1) through (3) for all inputs.
% \item Pick the input with greatest difference between expected error and current model expected error.
% \end{enumerate}
%An unlabeled data point is selected and assigned one of the possible labels.
%This ``fake'' data point is used to train a new objective function model and the error of that model is calculated over all remaining unlabeled data points.
%Error is combined across possible labels for each unlabeled data point.
%Expected error reduction then picks the unlabeled point with the least average (``expected'') error.
\item \textbf{Variance Reduction} Same as expected error reduction, but seeks to reduce variance in output predictions rather than error \cite{settles2012:al-book}.
% Corresponds to picking designs that lead to greatest reduction in uncertainty about the design space over time.
\end{itemize}

%%%%%%%%%%%%%%%%%%%%%%%%%%%%%%%%%%%%%

\section{Game Domain}

\begin{figure}[t]
\centering
\includegraphics[width=1\linewidth]{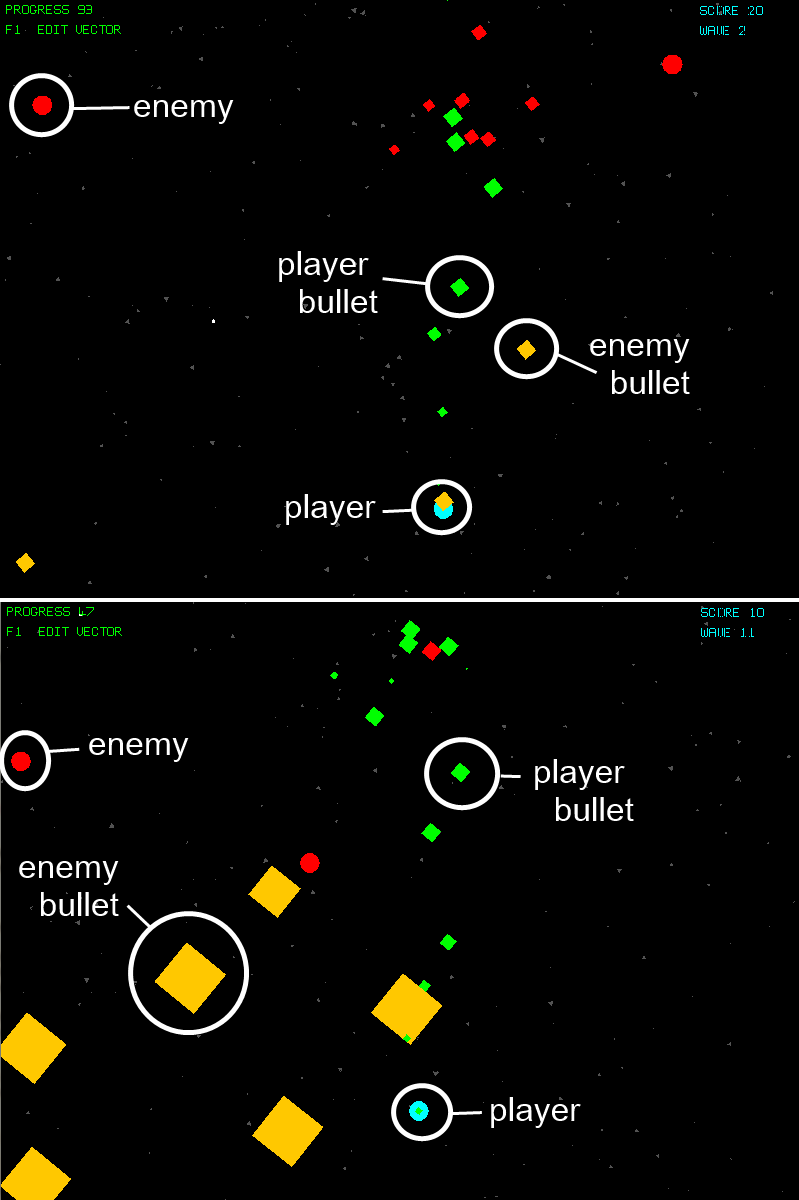}
\caption{Study game interface illustrating player, enemies, and shots fired by both at two points along adaptation process.}
\label{fig:shmup}
\end{figure}

We sought to assess how well AL could reduce the number of playtests needed to achieve a design goal.
To conduct a case study of machine--driven playtesting we developed a simple shoot-`em-up game (Figure~\ref{fig:shmup}).
Shoot-`em-up games emphasize reflexes and pattern recognition abilities as a player maneuvers a ship to dodge enemy shots and return fire.
In general, arcade games serve as an ideal starting domain for low-level parameter tuning:
\begin{itemize}
\item There are a number of parameters that can potentially interfere with each other: size and speed of enemies and enemy bullets, rate of enemy fire, player speed, player rate of fire, etc.
\item The game can be played in a series of waves, enabling our system to naturally test game parameter settings and gather player feedback.
\item Action-oriented gameplay reduces the complexity of player long-term planning and strategizing.
\item A scoring system makes gameplay goals and progress clear, unlike domains involving puzzle-solving or aesthetic enjoyment of a game world or setting.
\end{itemize}
In the case of shoot-`em-up games, we tested two different kinds of game design goals: (a) player game play behavior goals and (b) player subjective response goals.
Player game play behavior goals cover cases where designers desire particular play patterns or outcomes---e.g. player success rates or score achieved. 
Subjective responses goals cover cases where designers desire specific player subjective feedback---e.g. getting good user ratings on the feel of the controls.

% % enemy params
The shoot-`em-up game involves space ship combat over a series of waves.
During each wave a series of enemies appear that fire bullets at the player. 
To test AL for regression we set a game play behavior design goal (objective function) of the player being hit exactly six times during each wave of enemies (output) and tuned enemy parameters (input).
We varied the size of enemy bullets, speed of enemy bullets, and rate that enemies fire bullets. 
Increasing bullet size requires the player to move more carefully to avoid bullets. 
Faster bullets require quicker player reflexes to dodge incoming fire. 
More rapid firing rates increase the volume of incoming fire. 
Together these three parameters govern how much players must move to dodge enemy attacks, in turn challenging player reflexes. 
Getting approximate settings for these parameters is easy, but fine-tuning them for a desired level of difficulty can be challenging.

% % control params
To test AL for classification we set a subjective response design goal (objective function) of the player evaluating a set of controls as better than the previous set (output) and tuned player control parameters (input).
We varied two ship movement parameters: drag and thrust. 
Drag is the ``friction'' applied to a ship that decelerates the moving ship at a constant rate when it is moving---larger values cause the ship to stop drifting in motion sooner. 
Thrust is the ``force'' a player movement press applies to accelerate the ship---larger values cause the ship to move more rapidly when the player presses a key to move. 
Combinations of thrust and drag are easy to tune to rough ranges of playability.
However, the precise values needed to ensure the player has the appropriate controls are difficult to find as player movement depends on how enemies attack and individual player preferences for control sensitivity (much like mouse movement sensitivity).
After each wave of enemies a menu asked players to indicate if the most recent controls were better, worse, or as good/bad as (``neither'') the previous set of controls.
We provided a fourth option of ``no difference'' for when players could not distinguish the sets of controls, as opposed to ``neither'' where players felt controls differed but had no impact on their preferences.

%Our experiments investigate how well an AI system can learn the right parameter settings to achieve a desired level of player performance (in terms of rate of being hit by enemies) or a most preferred set of controls (in terms of player subjective responses).

%%%%%%%%%%%%%%%%%%%%%%%%%%%%%%%%%%

\section{Experiments}
Our experiments tested whether AL could reduce the number of human playtests needed to tune design parameters compared to a random sampling approach.
Random sampling is the standard baseline used to evaluate the efficacy of AL models for improving an objective function for a fixed number of inputs \cite{settles2012:al-book}. 
Random sampling is similar to A/B testing approaches that capture large amounts of data before acting on the results.
Randomizing parameter ordering helps control for player learning over waves.

In two experiments we used the above AL acquisition functions for regression by tuning enemy parameters and for classification by tuning player controls, respectively.
For both experiments we first built a data set by providing human players with random sets of parameters and recording behavior or subjective responses, respectively.
The experiments had AL methods use this data as a pool of potential playtests to run and evaluated how well AL could pick a sequence of playtests to best achieve the design goals.
%Both experiments involved a preliminary test on simulated data followed by testing with human participants.
%The simulation experiments allowed us to verify that our methods would work in principle; humans studies show our method can apply to real-world contexts.

In the regression study we used Gaussian Processes (GPs), the standard non-linear regression function in the Bayesian experimental design literature.
GPs generally yield good models with few playtests (samples) and have computationally inexpensive analytic formulations for many of our acquisition functions.
In the classification study we used three different objective functions---Gaussian Processes (GP), Kernel Support Vector Machines (KSVM), and optimized neural networks (``neuro-evolution'', NE).
KSVMs and NE are common classification approaches, whereas GPs are not.
Kernel methods (e.g. KSVMs and GPs) are a popular machine learning technique previously used in player modeling \cite{yu2011:minboredom} and optimized neural networks have been widely used in preference learning \cite{yannakakis2011:edpcg}.\footnote{For computational reasons we use a gradient-based optimization method for network structure, size, and weights, rather than the more common neuro-evolutionary approaches. 
We found no performance differences between the two optimization approaches in initial tests on our data.}
%All three of these objective functions can be trained in regression or classification modes.
Since NE does not produce probabilistic predictions it cannot use some of the above acquisition functions.

\subsection{Data Collection}

We empirically evaluated AL by deploying two versions of our game online.
We publicized the game through websites and local emailing lists and did not offer compensation to participants.
To collect data on patterns of play over time we asked participants to try to play at least 10 waves of the game, though we did not enforce this requirement.

For analysis we only used data from players who played at least 10 waves total.
This ensures we avoid data from players who were unable to reliably run the game.
% AZ 2013/10/29 - planning to rerun w/requirement of min 10 waves total, only waves 1-10
%	AZ 2013/11/27 - done
For our regression experiment this resulted in data from 138 players and 991 waves of the game total (using all waves each player played).
For our preference experiment we had 57 players, 47 of these provided only binary responses of ``better'' or ``worse'' and we limited our analysis to this subset of players to yield 416 paired comparisons.
% AZ 2013/10/29 - planning to rerun w/requirement of min 10 waves total; should filter some of the people with little data and thus more likely to be spurious results
%	AZ 2013/11/27 - done
We only used preference responses during the first 10 waves of play to avoid collecting many positive responses from those who were highly engaged in the game.
Note that we did not collect preference comparisons for the first wave of the game as players could not yet compare control settings.

\subsection{Experiment Design}

% experiment design
Using this data we performed 10-fold cross-validated experiments to measure how well a playtesting strategy (acquisition function) could achieve a design goal (objective function) given a set of design parameters (input).
% regression
For regression we trained a GP (design model) using the three enemy parameters (input: bullet speed, bullet size, and firing rate) to minimize the squared difference between the number of times the player was hit (output) and a desired rate of 6 times (objective function).
We squared the difference to more steeply penalize sets of parameters with greater differences from the ideal.
% classification
For classification we trained a GP, KSVM, or NE (design model) with four control parameters (input: current and previous drag and thrust) to predict player control preference comparisons as ``better'' or ``worse'' (output) with a design goal of maximizing prediction quality (objective function: F1 score).
%to maximize the prediction quality (as F1 score) on whether the preference rating was ``better'' or ``worse'' (objective function).
%We discarded ratings that were not in these two classes as our data had too few samples to make a comparison (only 10/57 players ever responded in other categories).

For each cross-validation fold we first randomly selected 10\% of the data and set it aside for evaluating objective function performance.
%randomly selected 300 data points to use for cross-validation; the remaining data was used as .
Next we randomly sampled 30 input--output pairs from the other 90\% of the data set to create a training data set; the remaining unused samples formed the training pool.
%An additional set of 500 testing points was generated in the same way.
%We then created the grid of points the AL model could sample, forming the training pool.
Within each fold we then repeated the following process:
\begin{enumerate}
\item Train the regression or classification model on the training data set.
\item Evaluate the objective function for that model on the testing data set.
\item Use the acquisition function to pick a new input sample from the training pool (without yet knowing the sample output) to improve the objective function.
\item Move the selected sample (including the true output) from the training pool into the training data.
\item Return to the first step and repeat the process until the maximum number of training samples are used.
\end{enumerate}
\noindent We used a maximum of 300 training samples in both regression and classification.

% regression: 138 players (currently using max 10 waves verion)
%% note: currently kept up to 10 waves, but allowed fewer
%% if min 10 waves -> 82 players, 1753 data points
%% if max 10 waves -> 138 players, 991 data points [current]
%% if both -> 82 players; 819 data points
% preference: 57 players (in random) with any value; 47 with binary choices
%% if binary but no wave -> 47 players, 318 points [current]
%% if binary min 10 waves -> 41 players, 296 points

%%%%%%%%%%%%%%%%%%%%%%%%%%%%%%%%%%

\section{Results and Discussion}

Overall our results show AL is a promising approach for reducing the number of playtests needed to achieve a design goal.
%In regression and classification studies AL was able to reduce the number of playtests using multiple different acquisition functions.
For enemy parameter tuning (a regression problem) we found acquisition functions that balance exploration and exploitation (especially UCB) have the best performance.
For control tuning (a classification problem) we found acquisition functions that tolerate high variance (e.g. QBB and entropy) have strong performance.
No single acquisition function, objective function, or acquisition-objective function pair was optimal across cases and number of playtests.
These results align with previous work in AL showing that many data-specific properties impact AL efficacy \cite{schein2007:al-logreg-eval}.
Below we provide further details with an emphasis on how AL impacted the need for playtesting.

\subsection{Regression}

Our regression experiments show AL can effectively tune for players having a desired level of performance, even with few samples.
Having a clear behavioral objective (being hit a number of times in the game) was likely a strong contributor.
We found UCB was most effective (Table~\ref{tab:reg_res}).\footnote{AL methods led to significant improvements over the random baseline in all reported results tables.}
UCB explicitly balances exploring and exploiting potential test designs, suggesting parameter tuning objectives involve a balance between considering alternative parameter settings and refining a given setting.

\begin{figure}[tb]
\centering
\includegraphics[width=\linewidth]{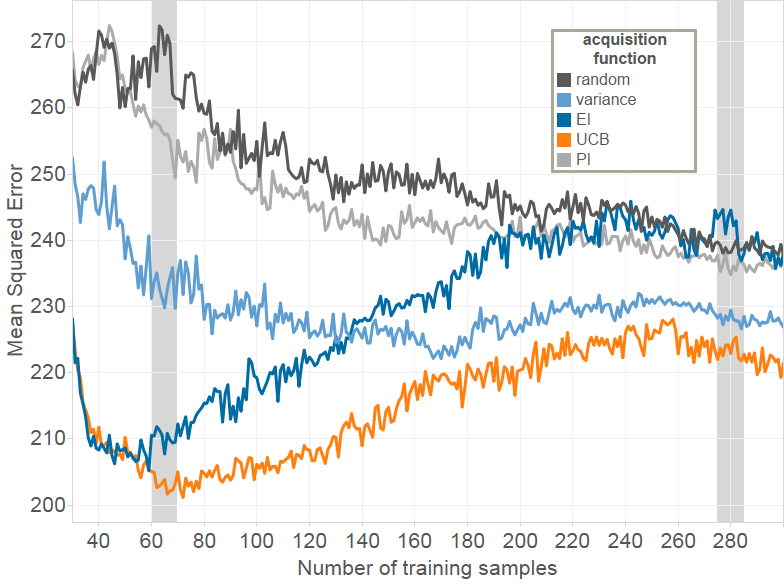}
\caption{GP performance using different acquisition functions.
Shows MSE with an increasing pool of AL-selected training samples.
Lower values indicate better performance.
Bands indicate values that were averaged to produce Table~\ref{tab:reg_res}.
}
\label{fig:reg_happy}
\end{figure}

\begin{figure}[tb]
\centering
\includegraphics[width=\linewidth]{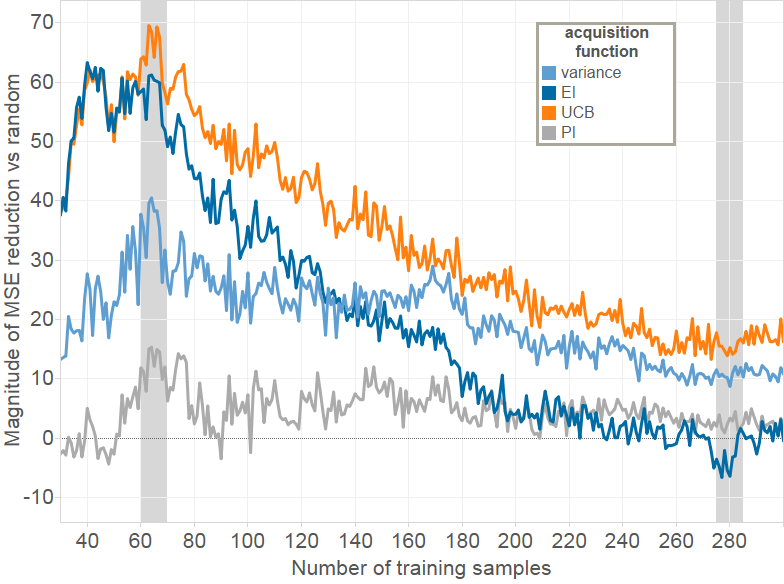}
\caption{GP performance improvement over random sampling using different acquisition functions.
Shows amount of MSE reduction with an increasing pool of AL-selected training samples.
Larger values indicate better performance.
}
\label{fig:reg_all}
\end{figure}

%% human
% UCB and EI are only ones to do well
% -- UCB remains at consistently high performance
% -- EI gradually gets worse over time
% why?
% -- variance and PI do not capture exploration and thus have difficulty finding good results
% -- UCB is able to temper exploration and exploitation over time, likely helping focus on best areas to test
AL methods that do not balance exploration and exploitation had worse performance either with few or many samples.
Figure~\ref{fig:reg_happy} shows MSE values (higher is worse performance) for model predictions when trained with different acquisition functions, illustrating the better performance of all AL methods over random sampling; Table~\ref{tab:reg_res} provides values at selected regions.
Figure~\ref{fig:reg_all} shows how much different acquisition functions reduced MSE compared to a baseline random sampling approach (larger values indicate greater improvements).
From these figures it is clear AL is particularly advantageous at small sample sizes, though most methods show improvements up to the maximum number of samples used.

Variance performed relatively better with many samples, explained by the need to explore heavily before having a good enough design model.
When tuning many parameters at once it is easy to find many sets of uncertain (but bad) parameters, leading to poor performance with few samples.
Over time EI gradually worsened while UCB and variance maintained better performance.
As more samples are gathered UCB reduces exploration while EI eventually begins to make poor playtest choices.
Approximately 70 samples were needed to train the successful AL methods for the largest peak performance improvements; random sampling never achieved this level of performance on our data set (Table~\ref{tab:reg_res}).
Overall this clearly demonstrates AL can enhance playtesting efficacy, perhaps beyond what would happen through simply A/B testing and collecting data as suggested by the asymptotically higher performance of UCB and variance.

% summary
Our regression experiments show the power of AL to reduce the amount of playtesting required and better achieve design goals.
UCB's balance of exploration and exploitation had the greatest efficacy and suggests a gradual refinement design process is optimal.
These results make a strong case for AL applied to optimizing low-level in-game behaviors, such as difficulty in terms of in-game performance.

% http://www.tablesgenerator.com/

\begin{table}[tb]
\caption{Regression GP MSE acquisition function comparison.
Sample sizes indicate values averaged over a range of $\pm$5 samples (for smoothing).
Lower values indicate better performance.}
\centering
\begin{tabular}{|c|c|c|}
\hline
acquisition function & 65 samples   & 280 samples  \\ \hline
random               & 268          & 239          \\ \hline
variance             & 233          & \textbf{228} \\ \hline
PI                   & 255          & 236          \\ \hline
EI                   & \textbf{210} & 242          \\ \hline
UCB                  & \textbf{203} & \textbf{224} \\ \hline
\end{tabular}
\label{tab:reg_res}
\end{table}

\subsection{Classification}
Our classification experiments show AL improves models of subjective player preferences with both probabilistic and non-probabilistic acquisition functions.
Methods that tolerate high variance---entropy, QBB vote and probability, and expected error reduction---have the strongest performance (Table~\ref{tab:cls_res}).
These acquisition functions succeed by overcoming the noise inherent in human playtest data, particularly when using few playtests.
Performance increases with few playtests means AL may be particularly effective in cases where playtesting time or resource budget is limited.
Our results show AL methods are effective even with more complex data and can improve a variety of baseline classification design models (called classifiers---GPs, KSVMs, and NE).

\begin{figure}[tb]
\centering
\includegraphics[width=\linewidth]{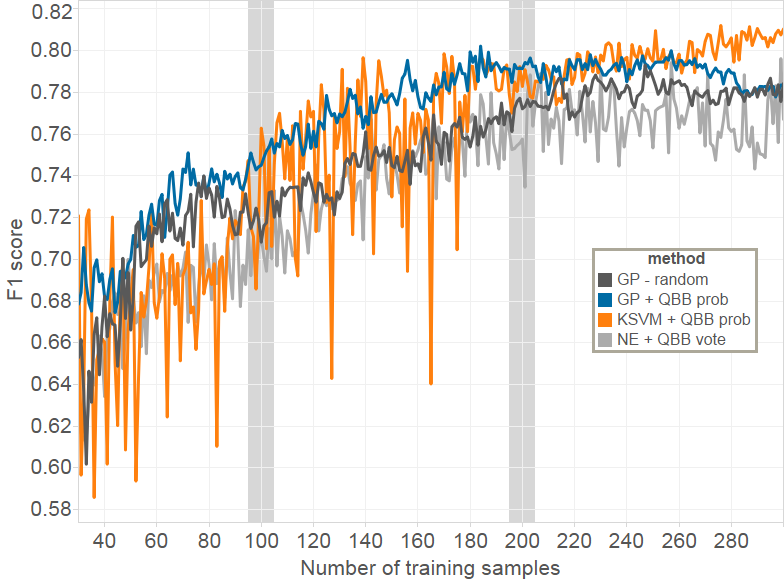}
\caption{Classification performance with different combinations of classifiers and acquisition functions.
Higher values indicate better performance.
Shows F1 score with an increasing pool of AL-selected training samples.
Bands indicate values that were averaged to produce Table~\ref{tab:cls_res}.
Only the best-performing acquisition functions for each classifier are shown for clarity.
%The only random sampling shown is using a GP as this had the best performance.
}
\label{fig:cls_happy}
\end{figure}

\begin{figure}[tb]
\centering
\includegraphics[width=\linewidth]{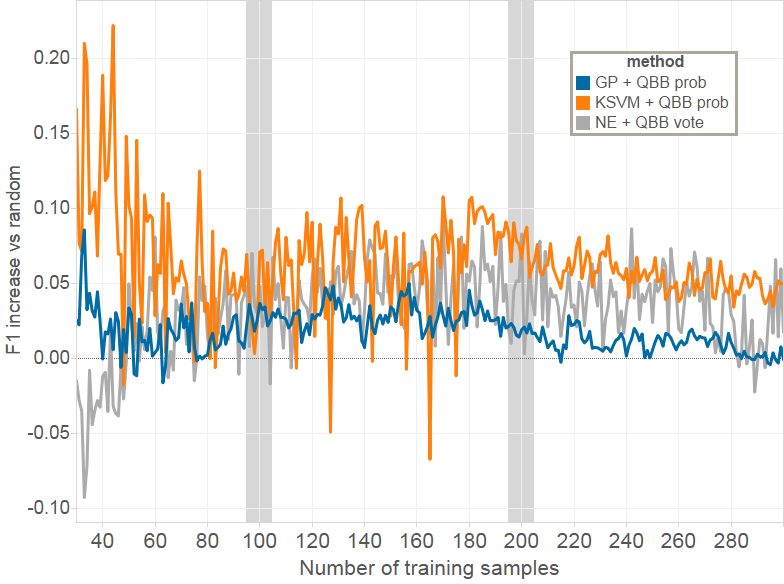}
\caption{Classification performance improvement over random sampling with different combinations of classifiers and acquisition functions.
Higher values indicate better performance.
Shows gains in F1 score with an increasing pool of AL-selected training samples.
Only the best-performing acquisition functions for each classifier are shown for clarity.
}
\label{fig:cls_all}
\end{figure}

% human:
%	random: GP > NE > KSVM
%	AL: GP random/QBB prob > NE QBB vote > KSVM QBB prob
%		GP: QBB prob +vote wins w/few samples, loses/ties at more
%		NE: no difference
%		KSVM: QBB vote early improvement
Entropy, QBB vote and probability, and error reduction all improved classification quality (as F1 score) over random sampling.
%Figure~\ref{fig:cls_all} shows the amount of performance improvement for the best-improving acquisition functions for each classifier; Table \ref{tab:cls_res} shows the raw F1 scores.
Figure~\ref{fig:cls_happy} shows F1 scores (higher is better performance) for the best performing acquisition functions for each classifier; Table~\ref{tab:cls_res} provides values at selected regions.
Figure~\ref{fig:cls_all} shows how much different acquisition functions increased F1 scores compared to a baseline random sampling approach using the same classifier (larger values indicate greater improvements).
These figures illustrate AL can provide substantial gains with few samples and maintain an improvement over random sampling up to the maximum number of samples used.

QBB methods (especially vote) were effective at both few and many samples.
Entropy was only effective with few samples while error reduction was most effective with more samples.
Expected error reduction must predict future outcomes and thus requires more initial data before becoming effective.
Variance reduction had poor performance.
As with the variance acquisition function for regression a large number of possible parameters causes difficulty in effectively reducing variability in responses.
We speculate preference responses are typically noisy due to people shifting preferences or disagreeing on a common design as preferable (e.g. mouse control sensitivity in first-person games).

% base model comparison
Comparing the design models, we found GPs had the best baseline performance (with random sampling), followed by NE and then KSVMs.
Overall GPs with QBB probability or expected error reduction did best, followed by KSVMs with either QBB method and then NEs using QBB vote.
Using AL methods provided the largest performance boost for KSVMs, though GPs and NE also benefited.

% summary
Our classification experiments thus demonstrate AL can reduce the amount of playtesting needed even for subjective features of a design such as control settings.
Reducing playtest costs requires acquisition functions (e.g. entropy, QBB, and error reduction) that mitigate the noise inherent in preference response data.
AL always improved over random sampling across different design model approaches, though the best acquisition functions varied.

\begin{table}[tbp]
\centering
\caption{Classification acquisition-objective function F1 score comparison.
Sample sizes indicate values averaged over a range of $\pm$5 samples (for smoothing).
Higher values indicate better performance.}
% % scriptsize version
\scriptsize
\begin{tabular}{|c|c|c|c|c|c|c|}
\hline
\multirow{2}{*}{\begin{tabular}[c]{@{}c@{}}acquisition\\ function\end{tabular}} & \multicolumn{3}{|c|}{100 samples}                & \multicolumn{3}{|c|}{200 samples}                \\ \cline{2-7} 
\multicolumn{1}{|l|}{}                                                          & GP             & KSVM           & NE             & GP             & KSVM           & NE             \\ \hline
random                                                                          & 0.720          & 0.684          & 0.673          & 0.773          & 0.709          & 0.718          \\ \hline
entropy                                                                         & \textbf{0.763} & \textbf{0.731} & N/A            & 0.763          & 0.751          & N/A            \\ \hline
QBB vote                                                                        & \textbf{0.758} & \textbf{0.746} & \textbf{0.703} & 0.780          & \textbf{0.777} & \textbf{0.760} \\ \hline
QBB prob                                                                        & 0.749          & 0.724          & N/A            & \textbf{0.792} & \textbf{0.782} & N/A            \\ \hline
error red                                                                       & \textbf{0.761} & 0.702          & N/A            & \textbf{0.795} & \textbf{0.772} & N/A            \\ \hline
var red                                                                         & 0.660          & 0.667          & N/A            & 0.725          & 0.723          & N/A            \\ \hline
\end{tabular}

\label{tab:cls_res}
\end{table}

\subsection{Limitations}

Our work has several limitations that point to important avenues for further developing machine--driven playtesting.
We used a game domain that allows for a tight loop between playing the game and evaluating the game parameters.
More complex design tasks such as playtesting a platformer level or tuning a simulation game system will likely require more sophisticated techniques for credit assignment to design elements.
Our tasks tuned a set of flat parameters.
Tasks with structured design elements---e.g. rule sets in a card or board game---will require alternative techniques able to handle modular composition of design parameters in the optimization process.
Understanding the relative strengths and weaknesses of different AL approaches for different game types and design tasks is a rich area for further research.
This paper presents a first step toward understanding how machines can support game design across these applications.

%%%%%%%%%%%%%%%%%%%%%%%%%%%%%%%%%%

\section{Conclusions}

% % note: can combine multiple aspects into single objective function, but then need additional complexity of learning how to balance objectives if not told directly; e.g. preference + performance

% % enable continuous model improvement

% % future: optimal experimental design (rafferty, chaloner); hcomp for more direct human participation

% value: show designers how best to playtest!

We have shown how playtesting for low-level design parameter tuning can be automated using active learning.
AL can reduce the cost of playtesting to achieve a design goal by intelligently picking designs to test.
In some cases AL may get better results for a design goal than simple A/B testing could accomplish.
AL is especially powerful when coupled to online playtesting, enabling a machine to simultaneously optimize many parameters without burdening the designer.
%machine--driven playtesting has great promise for broadening how automation enhances existing game design practices across the design iteration process.

To the best of our knowledge this paper is novel in automatically tuning game controls. %, specifically meeting player preferences.
Application of machine learning and AI to games have focused on content generation rather than player \textit{interaction}.
Interaction ``feel'' is crucial to many digital games---poor controls can obstruct player enjoyment or make a game unplayable.
This work provides a first step to addressing this problem through tuning controls to construct a single control setting that best meets the preferences of many players (e.g. default controls).

We believe machine--driven playtesting can provide insights into how to best perform playtests.
Designers might learn better playtest techniques by considering how machines best perform these tasks.
For example, UCB was the best acquisition function for AL in regression.
UCB uses a process of initially exploring widely before shifting toward greater exploitation by choosing designs expected to elicit both high-quality and high-variability results to narrow the space of alternatives.
Balance or other feature--tuning techniques may benefit from similar methods of testing multiple alternatives based on balanced consideration of their value and how well they are understood.

We anticipate AL can improve game development processes by making playtesting more efficient and cost-effective.
AL may also inform processes that cannot yet be automated: e.g. using forums or bug reports to identify and prioritize potential design changes.
Machine--driven playtesting can thus complement the strengths of human game designers by allowing them to focus on high-level design goals while machines handle mundane parameter tweaking.

%%%%%%%%%%%%%%%%%%%%%%%%%%%%%%%%%%

%\section{Acknowledgments}
% Eric Fruchter - or possible co-author
% NSF grant

\pagebreak

\bibliographystyle{abbrv}
\bibliography{lib}  % keeping lib in separate github project

\end{document}